\documentclass[conference]{IEEEtran}
\IEEEoverridecommandlockouts

\usepackage{cite}
\usepackage{amsmath,amssymb,amsfonts}
\usepackage{algorithmic}
\usepackage{graphicx}
\usepackage{textcomp}
\usepackage{xcolor}
\usepackage{multirow}
\usepackage{bbding}
\usepackage{soul}
\def\BibTeX{{\rm B\kern-.05em{\sc i\kern-.025em b}\kern-.08em
    T\kern-.1667em\lower.7ex\hbox{E}\kern-.125emX}}
\begin{document}

\title{MSECG: Incorporating Mamba for Robust and Efficient ECG Super-Resolution\\
}

 
\author{\IEEEauthorblockN{
Jie Lin\textsuperscript{*}\IEEEauthorrefmark{2},
I Chiu\textsuperscript{*}\IEEEauthorrefmark{5},
Kuan-Chen Wang\textsuperscript{*}\IEEEauthorrefmark{3},
Kai-Chun Liu\IEEEauthorrefmark{4},
Hsin-Min Wang\IEEEauthorrefmark{5},
Ping-Cheng Yeh\IEEEauthorrefmark{3},
and Yu Tsao\IEEEauthorrefmark{5}}

\IEEEauthorblockA{
\IEEEauthorrefmark{2}Department of Information Management, National Taiwan University, Taiwan
\IEEEauthorrefmark{5}Academia Sinica, Taiwan \\
\IEEEauthorrefmark{3}Graduate Institute of Communication Engineering, National Taiwan University, Taiwan \\
\IEEEauthorrefmark{4}College of Information and Computer Sciences, University of Massachusetts, Amherst, USA\\
Email: b11705048@ntu.edu.tw,\ 
d13949002@ntu.edu.tw,\ 
d12942016@ntu.edu.tw,\ 
kaichunliu@umass.edu,\\
whmat@iis.sinica.edu.tw,\ pcyeh@ntu.edu.tw,\ yu.tsao@citi.sinica.edu.tw
}

}

\maketitle

\begingroup
\renewcommand\thefootnote{\textsuperscript{*}}
\footnotetext{These authors contributed equally to this work.}
\endgroup

\begin{abstract}
Electrocardiogram (ECG) signals play a crucial role in diagnosing cardiovascular diseases. To reduce power consumption in wearable or portable devices used for long-term ECG monitoring, super-resolution (SR) techniques have been developed, enabling these devices to collect and transmit signals at a lower sampling rate. In this study, we propose MSECG, a compact neural network model designed for ECG SR. MSECG combines the strength of the recurrent Mamba model with convolutional layers to capture both local and global dependencies in ECG waveforms, allowing for the effective reconstruction of high-resolution signals. We also assess the model's performance in real-world noisy conditions by utilizing ECG data from the PTB-XL database and noise data from the MIT-BIH Noise Stress Test Database. Experimental results show that MSECG outperforms two contemporary ECG SR models under both clean and noisy conditions while using fewer parameters, offering a more powerful and robust solution for long-term ECG monitoring applications.
\end{abstract}


\section{Introduction}
\label{sec:intro}
Cardiovascular diseases pose significant life-threatening risks, often preventable with early detection of Cardiac Arrhythmias (CAs) \cite{berkaya2018survey}. Electrocardiogram (ECG) serves as a non-invasive diagnostic tool, measuring the heart's electrical activity and aiding in monitoring cardiac anomalies \cite{hurst1998naming}. To facilitate the detection of CAs in daily life, wearable or portable devices are commonly employed for long-term ECG monitoring \cite{huda2020low, xia2018automatic}. Reducing power consumption in these devices is crucial, and this can be achieved by collecting and transmitting ECG with a low sampling rate~\cite{nishikawa2018sampling}. However, lowering the sampling rate risks compromising diagnostic accuracy due to information loss \cite{pizzuti1985digital}, and traditional upsampling methods often produce severe artifacts that further hinder accurate diagnosis \cite{krylov2009combined,chen2023srecg}.

Super-resolution (SR) techniques aim to recover high-resolution (HR) data from low-resolution (LR) versions. Modern SR methods, leveraging neural networks (NNs), have found extensive applications in fields such as computer vision \cite{dong2015image,ledig2017photo} and audio processing \cite{yoneyama2023nonparallel,yu2023conditioning}. In ECG analysis, NN-based SR methods have emerged, such as SRECG \cite{chen2023srecg}, which applies a convolutional neural network (CNN) inspired by SRResNet \cite{ledig2017photo} to reconstruct HR ECG from LR inputs. Similarly, DCAE-SR \cite{lomoio2024dcae} utilizes a denoising autoencoder structure for ECG SR under noisy conditions. These studies demonstrate the feasibility of using NNs to recover HR ECG signals from low-sampling-rate inputs. However, existing approaches based on convolutional structures may be limited in capturing temporal dependencies inherent in data~\cite{vaswani2017attention}, leading to constrained ECG reconstruction capability.

Mamba \cite{gu2023mamba}, a recently developed recurrent neural network (RNN), enhances traditional state-space models (SSMs) by introducing a selection mechanism and an efficient hardware-aware algorithm. Compared to Transformer~\cite{vaswani2017attention}, which also handles global dependencies but with higher time complexity, Mamba offers a more efficient solution. Owning to its advantages, Mamba has been applied across diverse fields like speech processing \cite{chao2024investigation}, computer vision \cite{zhu2024vision,ma2024u}, and genomics \cite{zhang2024chimamba, thoutammsamamba}. However, its potential for ECG SR has not yet been explored, as most existing approaches rely on Convolutional Neural Networks (CNNs).

In this study, we propose MSECG, a novel Mamba-based ECG SR model for reconstructing HR ECG with enhanced accuracy. MSECG combines convolutional layers with Mamba to capture both local and global information within data. Additionally, advanced SR techniques, such as pixel shuffle (PS) and skip connection (SC) \cite{shi2016real}, are incorporated to further boost performance and efficiency. Our experimental results show that MSECG outperforms contemporary methods on the PTB-XL ECG database, even under noisy conditions using data from the MIT-BIH Noise Stress Test Database (NSTDB). These findings underline MSECG’s potential as a robust and effective SR solution for real-world ECG applications. To the best of our knowledge, this is the first work to explore Mamba's application in ECG SR.

\begin{figure*}[htb]
    \centering
    \includegraphics[width=0.8\textwidth]{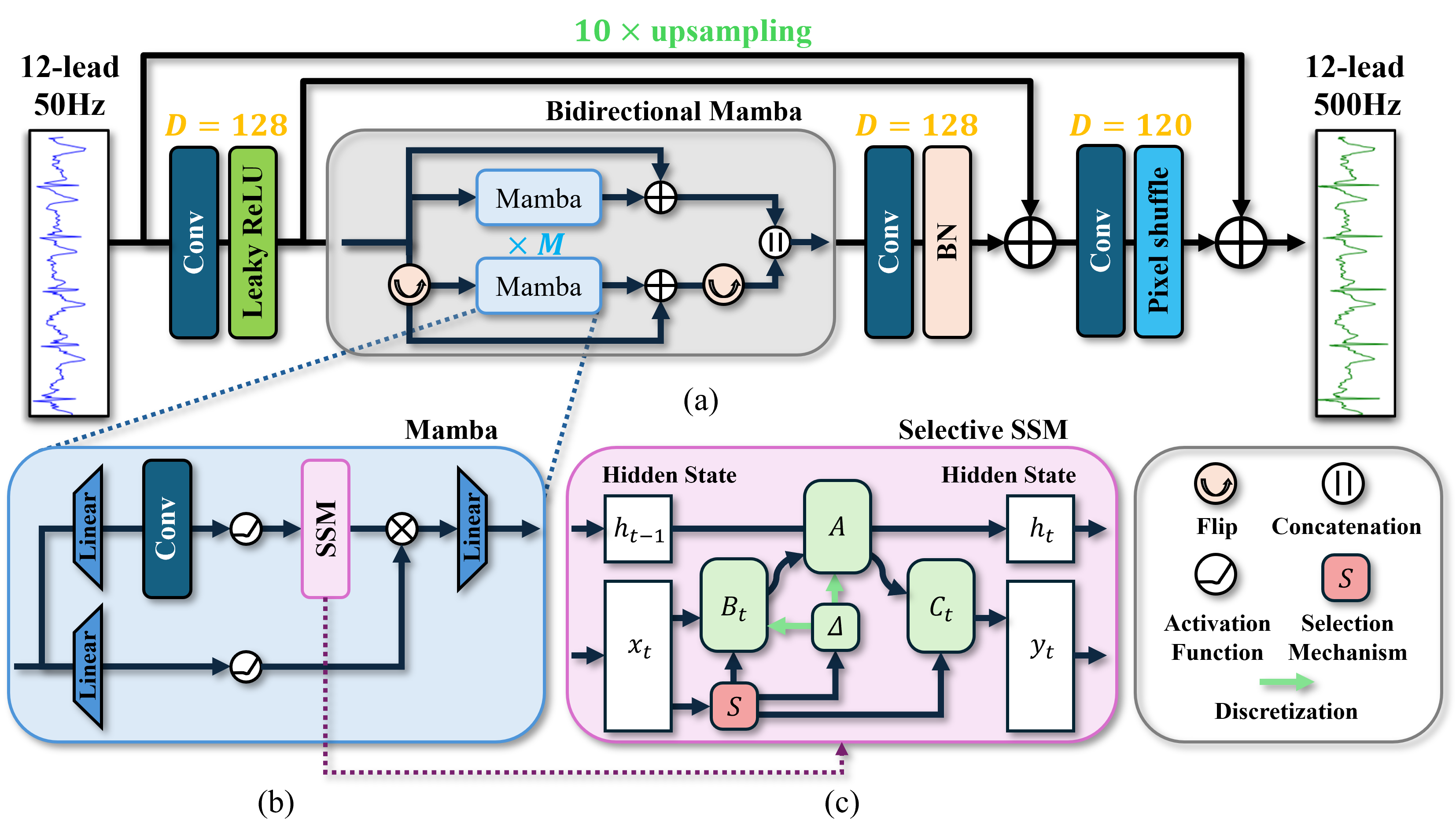}
    \caption{The architecture of (a) MSECG, (b) Mamba, and (c) selective SSM.}
    \label{fig:msecg}
\end{figure*}

\section{Related Works}\label{sec:related}

\subsection{ECG Super-Resolution}\label{ssec:ecg_sr}
Reducing the power consumption of wearable and portable devices is essential for long-term ECG monitoring. To address this, many studies have focused on developing SR techniques for ECG data, aiming to enable accurate analysis by collecting LR signals. One such approach is SRECG \cite{chen2023srecg}, which uses SRResNet \cite{ledig2017photo} as its foundation. It employs residual convolutional blocks to extract features from LR ECG signals, followed by two deconvolutional layers to upsample the signals and generate HR outputs. Another study, DCAE-SR \cite{lomoio2024dcae}, addresses the noise in ECG SR. DCAE-SR consists of an encoder and two decoders, each with four convolutional or deconvolutional layers. The encoder extracts latent features from the LR ECG signals, which are then fed into both decoders. One decoder reconstructs the original noisy LR ECG, while the other generates clean HR ECG signals. In our study, we implement both of these models to compare with our proposed method.

\subsection{Mamba}\label{ssec:mamba}
Mamba \cite{gu2023mamba} is an SSM designed for subquadratic-time computation. It addresses the limitations of Transformers \cite{vaswani2017attention}, whose computational and memory requirements grow quadratically as the input sequence length increases. Mamba introduces a selection mechanism that makes the model input-dependent, allowing it to focus on or ignore specific information as needed.
Mamba's architecture combines elements from the H3 model \cite{dao2023hungry} and gated multi-layer perceptron (MLP) blocks \cite{liu2021pay}, enabling it to be stacked homogeneously. This design expands the model's dimensionality, concentrating most parameters in linear projections, which enhances its ability to handle complex tasks.

In Mamba, structured SSMs map an input $\boldsymbol{x}$ to an output $\boldsymbol{y}$ through a higher-dimensional latent state $\boldsymbol{h}$, as described by the following equations: 
\begin{align}
    h_{n}&=\boldsymbol{\Bar{A}}h_{n-1}+\boldsymbol{\Bar{B}}x_{n},\\
    y_{n}&=\boldsymbol{C}h_{n},
\end{align}
where $\boldsymbol{\Bar{A}}$ and $\boldsymbol{\Bar{B}}$ denote discretized state metrices. The discretization process transforms "continuous parameters" $(\boldsymbol{\Delta}, \boldsymbol{A}, \boldsymbol{B})$ into "discrete parameters" $(\boldsymbol{\Bar{A}}, \boldsymbol{\Bar{B}})$, enabling the model to seamlessly handle discrete-time data.

Mamba also features a hardware-aware algorithm that scales linearly with the input sequence length, enabling faster recurrent computations through efficient scanning. Despite its concise structure, Mamba consistently achieves state-of-the-art performance in various domains, including speech processing \cite{chao2024investigation}, computer vision \cite{zhu2024vision,ma2024u}, and genomics \cite{zhang2024chimamba, thoutammsamamba}.

\section{Proposed method}\label{sec:material}

\subsection{Model architecture}
\label{ssec:model}
The architecture of the proposed MSECG is illustrated in Fig.~\ref{fig:msecg} (a). A convolutional layer initially extracts features from the input LR ECG. The features are then processed through a bidirectional Mamba block, retrieving information from both forward and backward directions in sequences. The structure of the Mamba block and selective SSM are provided in Fig.~\ref{fig:msecg} (b) and (c), respectively. We consider that the Mamba blocks can better capture temporal information than the residual convolutional blocks used in SRECG \cite{chen2023srecg}. The notations of $D$ and $M$ refer to the number of output channels in convolutional layers and the number of Mamba layers, respectively. 

\begin{figure}[tb]
    \centering
    \includegraphics[width=0.48\textwidth]{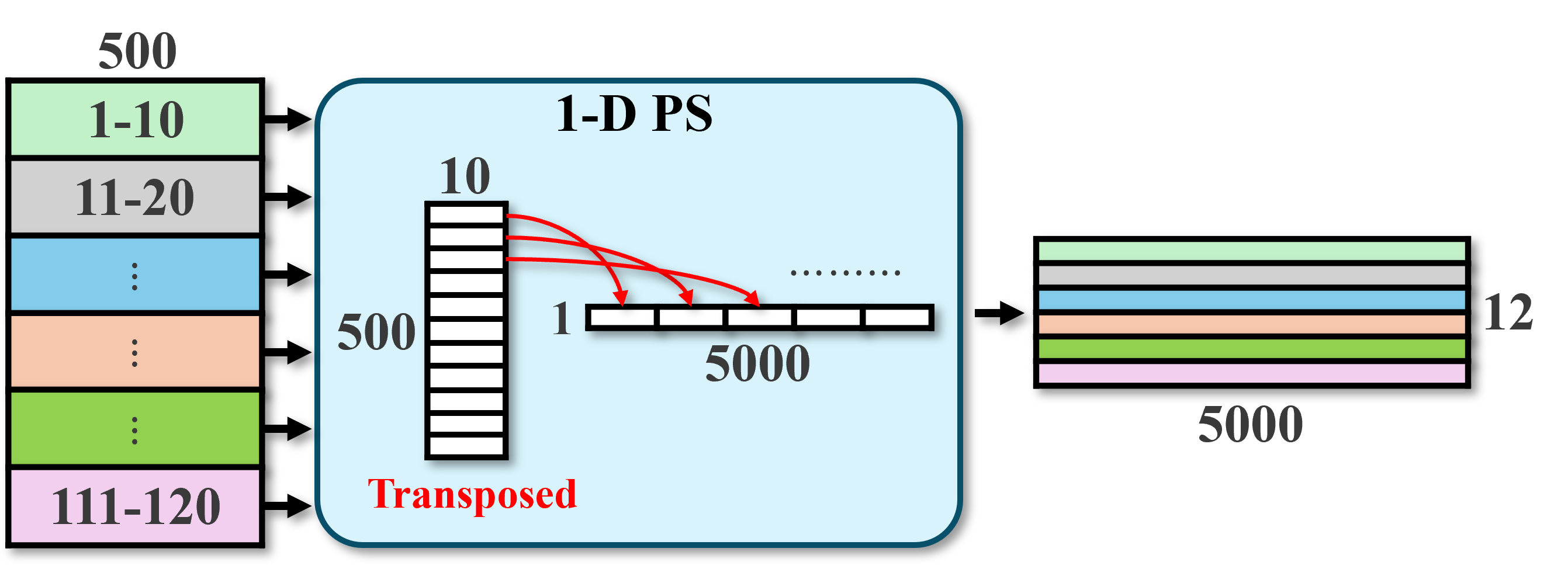}
    \caption{Illustration of one-dimensional pixel shuffling (PS) operation.}
    \label{fig:ps}
\end{figure}

For upsampling, MSECG employs a one-dimensional PS operation, which is more efficient than traditional deconvolutional layers \cite{sugawara2018super}. Fig.~\ref{fig:ps} illustrates the PS operation, where every ten channels in the feature maps are sequentially combined into a single output channel.
Moreover, since modeling the entire upsampling process is relatively complex, we deploy a skip connection (SC) from the model's input to the output, which is inspired by advanced super-resolution methods in computer vision \cite{conde2024deep}.
Specifically, the LR input ECG is upsampled by linear interpolation (LI) with a ratio of 10 and then added directly to the MSECG output as a residual connection.
This design can lead to more accurate reconstruction by enabling the model to focus on learning the difference between the LI-upsampled ECG and the target ECG.




\section{Experiments}
\label{sec:experiment}
\subsection{Datasets}
\label{ssec:dataset}

In this study, we use the PTB-XL dataset \cite{wagner2020ptb} as the source of clean ECG data. PTB-XL contains 21,799 clinical 12-lead ECG signals, each 10 seconds long, from 18,869 patients, with a sampling rate of 500 Hz. The dataset includes metadata for each ECG signal, such as weight, age, and sex, along with manual annotations for signal quality. PTB-XL is ideal for our experiments due to its large amount of high-quality ECG data.

For ECG noise data, we use the noise recordings from the MIT-BIH Noise Stress Test Database (NSTDB) \cite{moody1984noise} to emulate real-world noisy conditions, in contrast to the simulated noise used in the previous study \cite{lomoio2024dcae}. The NSTDB includes three common types of noise: baseline wander (BW), muscle artifact (MA), and electrode motion artifact (EM). These recordings were taken from volunteers using two channels at a sampling rate of 360 Hz, lasting approximately 30 minutes.



\subsection{Data pre-processing and preparation}
\label{ssec:preprocessing}
We apply a second-order Butterworth band-pass filter with cutoff frequencies of 1 and 45 Hz to the 500 Hz ECG data from the PTB-XL dataset, which helps extract the most valuable information from the ECG signals \cite{cao2023incepse}. The filtered ECG signals are treated as the ground truth (GT).
Following \cite{chen2023srecg}, which demonstrates a significant performance drop with a downsampling factor of 10, we downsample the GT signals from 500 Hz to 50 Hz by skipping every 9 sequential time units to generate the LR ECG signals.
From the PTB-XL standard, we use Folds 1-8 for training, Fold 9 for validation, and Fold 10 for testing.

To generate noisy ECG data, we downsample the noise signals from the MIT-BIH NSTDB from 360 Hz to 50 Hz to match the sampling rate of the LR ECG signals. Following the framework in \cite{lomoio2024dcae}, we introduce a 50\% probability of noise contamination for each ECG segment. If noise is added, each type of noise (BW, MA, and EM) is equally likely to be selected. We then assign a signal-to-noise ratio (SNR) for each noise type, chosen randomly from a reasonable range \cite{hu2024lightweight}. Finally, a random noise segment is selected, adjusted in amplitude, and added to the LR ECG signal to generate noisy data with the specified SNR.

\begin{table*}[htb]
\centering
\renewcommand\arraystretch{1.25}
\small
\caption{Overall performance of different ECG SR methods.}
\begin{tabular}{cccccc}
\hline\hline
Method & Param. (M) & MSE ($\times 10^{-3}$) $\downarrow$ & CoS ($\times 10^{-1}$) $\uparrow$ & SNR (dB) $\uparrow$ & MAD $\downarrow$ \\ \hline
LI          & -          & 7.477 $\pm$ 12.502 & 9.021 $\pm$ 0.895 & 8.592 $\pm$ 4.285  & 0.877 $\pm$ 0.403 \\
SRECG \cite{chen2023srecg} & 3.05  & 0.422 $\pm$ 0.485 & 9.937 $\pm$ 0.047 & 19.751 $\pm$ 2.595 & 0.371 $\pm$ 0.214 \\
DCAE-SR \cite{lomoio2024dcae} & 31.21 & 4.461 $\pm$ 13.472 & 9.756 $\pm$ 0.225 & 12.591 $\pm$ 2.894 & 0.962 $\pm$ 0.802 \\ \hline
MSECG (ours) & \textbf{1.91}  & \textbf{0.184} $\mathbf{\pm}$ \textbf{0.335} & \textbf{9.975} $\mathbf{\pm}$ \textbf{0.031} & \textbf{24.037} $\mathbf{\pm}$ \textbf{2.851} & \textbf{0.221} $\mathbf{\pm}$ \textbf{0.187} \\ \hline\hline
\multicolumn{6}{l}{\textbf{Bold} represents the best performance.}
\end{tabular}
\label{tab:overall}
\end{table*}

\subsection{Evaluation metrics}
\label{ssec:metric}
We use four metrics for signal quality to evaluate the performance of SR approaches \cite{lomoio2024dcae, li2023descod}. In the equations below, $\boldsymbol{s}$ denotes the predicted ECG waveform, $\boldsymbol{g}$ is the GT waveform, and $N$ is the number of samples in each segment.


$\bullet$ Mean Squared Error (MSE): Measures the average squared difference between $\boldsymbol{s}$ and $\boldsymbol{g}$.
\begin{equation}
    \mathrm{MSE} = \dfrac{1}{N}\sum_{n=1}^{N}(\boldsymbol{s}[n]-\boldsymbol{g}[n])^{2}.
\end{equation}


$\bullet$ Cosine Similarity (CoS): Measures the similarity between two vectors,  $\boldsymbol{s}$ and $\boldsymbol{g}$.
\begin{equation}
    \mathrm{CoS} = 
    \dfrac{\boldsymbol{s} \cdot \boldsymbol{g}}
    {\| \boldsymbol{s} \| \| \boldsymbol{g} \|},
\end{equation}
where $\| \cdot \|$ calculates the Euclidean norms of signals.

$\bullet$ SNR: Evaluates how well the signal can be distinguished from the noise by calculating the ratio of the power of $\boldsymbol{g}$ to the noise.
\begin{equation}
    \mathrm{SNR} = 10 \cdot \log_{10}\left(
    \dfrac{\sum_{n=1}^{N}\boldsymbol{g}[n]^{2}}
    {\sum_{n=1}^{N}(\boldsymbol{s}[n]-\boldsymbol{g}[n])^{2}}
    \right).
\end{equation}


$\bullet$ Maximum Absolute Distance (MAD): Measures the largest deviation between $\boldsymbol{s}$ and $\boldsymbol{g}$, which is critical for identifying short-period artifacts or noise.
\begin{equation}
    \mathrm{MAD} = \max|\boldsymbol{s}[n]-\boldsymbol{g}[n]|,\quad \mathrm{for}\  0 \leq n \leq N.
\end{equation}

Lower MSE and MAD values reflect better SR performance, and higher SSIM and SNR values indicate better SR performance.

\subsection{Implementation details}
\label{ssec:implement}
We use the Adam optimizer \cite{kingma2015adam} for training. The batch size is set to 64, and we apply the $L_{2}$ loss function. Training is done in two stages: the model is first trained for 300 epochs with a learning rate of 1e-4, and the best-performing model on the validation set is saved. In the second stage, the saved model is further trained for 50 epochs with the learning rate reduced to 1e-5.


\begin{table*}[htb!]
\centering
\renewcommand\arraystretch{1.25}
\small
\caption{Ablation study for the model structure of MSECG.}
\begin{tabular}{cccccccccc}
\hline\hline
Model & $M$ & Param. (M) & DConv & PS & SC & MSE ($\times 10^{-3}$) $\downarrow$ & CoS ($\times 10^{-1}$) $\uparrow$ & SNR (dB) $\uparrow$ & MAD $\downarrow$ \\ \hline
\multirow{6}{*}{MSECG} & 5 & 3.02 & \Checkmark & - & - & 0.206 & 9.972 & 23.508 & 0.228 \\
& 5 & 3.02 & \Checkmark & - & \Checkmark & 0.195 & 9.973 & 23.796 & 0.227 \\
\cline{2-10}
& 5 & 1.91 & - & \Checkmark & - & 0.263 & 9.963 & 22.292 & 0.320 \\
& 5 & 1.91 & - & \Checkmark & \Checkmark & \textbf{0.184} & \textbf{9.975} & \textbf{24.037} & \textbf{0.221} \\
\cline{2-10}
& 4 & 1.68 & - & \Checkmark & \Checkmark & 0.206 & 9.971 & 23.429 & 0.233 \\
& 6 & 2.14 & - & \Checkmark & \Checkmark & \underline{0.192} & \underline{9.974} & \underline{23.912} & \underline{0.222} \\
\hline\hline
\multicolumn{10}{l}{\textbf{Bold} and \underline{underline} represent the best and second best performance.}
\end{tabular}
\label{tab:ablation}
\end{table*}


\begin{table}[htb!]
\centering
\renewcommand\arraystretch{1.2}
\small
\caption{Performance under clean and noisy scenarios.}
\begin{tabular}{ccccc}
\hline\hline
Model & Data & MSE ($\times 10^{-3}$) $\downarrow$ & CoS $\uparrow$ & SNR (dB) $\uparrow$ \\ \hline
\multirow{2}{*}{SRECG} & clean & 0.400 & 0.994 & 19.880 \\
& noisy & 0.443 & 0.994 & 19.546 \\ \hline
\multirow{2}{*}{DCAE-SR} & clean & 4.265 & 0.977 & 11.304 \\
& noisy & 4.648 & 0.974 & 10.717 \\  \hline
MSECG & clean & 0.165 & 0.998 & 24.294 \\
(ours) & noisy & 0.203 & 0.997 & 23.726 \\
\hline\hline
\end{tabular}
\label{tab:clean_noisy}
\end{table}

\subsection{Results and discussion}
\label{ssec:result}
We compare the performance of MSECG with the conventional linear interpolation (LI) method and two NN-based ECG SR methods: SRECG \cite{chen2023srecg} and DCAE-SR \cite{lomoio2024dcae}. The overall performance of these methods is shown in Table~\ref{tab:overall}. MSECG outperforms the other methods across all four quality metrics, demonstrating its superior ability to capture temporal information essential for reconstructing HR ECG signals. In contrast, SRECG and DCAE-SR rely on convolutional structures, which are less effective at leveraging temporal information. 

Table~\ref{tab:ablation} presents an ablation study that explores the effectiveness of different components within MSECG architecture. The results show that MSECG achieves the best performance by using the PS operation along with SC, which directly combines the linearly upsampled input with the output. Replacing deconvolutional (DConv) layers with PS not only reduces the number of model parameters but also prevents checkerboard artifacts \cite{sugawara2018super} and other issues like ringing and aliasing \cite{krylov2009combined} associated with DConv and LI. The SC further improves the model performance by allowing the main network to focus on learning the residual information. Furthermore, we found that using five Mamba layers ($M=5$) strikes the best balance between performance and computational cost, which is why this configuration is used in MSECG.

\begin{figure}[hbt!]
    \centering
    \includegraphics[width=\columnwidth]{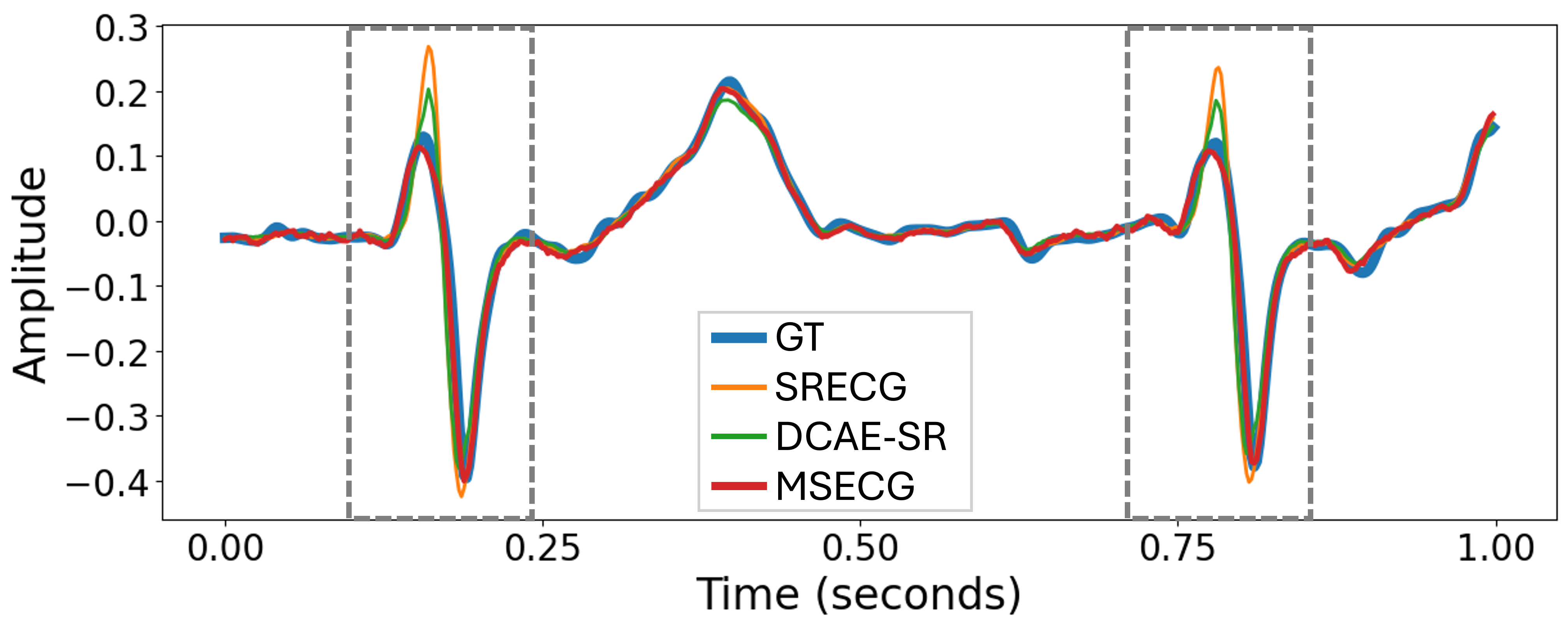}
    \caption{Reconstructed ECG signals using different SR methods. Dashed boxes highlight areas where the differences are most pronounced.}
    \label{fig:inference}
\end{figure}

To further assess the feasibility of ECG SR models in the real-world environment, Table~\ref{tab:clean_noisy} compares model performance in both clean and noisy conditions. Overall, noise degrades the performance across all metrics and models. However, MSECG continues to deliver the best results in both scenarios, demonstrating its robustness and potential for real-world ECG applications.

Fig.~\ref{fig:inference} shows the reconstructed HR ECG waveforms of different SR models. MSECG generates waveforms that are most similar to the GT signals, while both SRECG and DCAE-SR struggle to reconstruct the R-peaks accurately. Such distortion in ECG waveform and cardiac events could lead to misinterpretations in cardiovascular disease diagnosis~\cite{meng2021long,liu2018performance}.

\section{Conclusion}
\label{sec:conclusion}
In this study, we introduce MSECG, a novel ECG SR model that combines Mamba and convolutional layers to effectively capture both local and global information for HR ECG reconstruction. The experimental results demonstrate that MSECG achieves superior performance while using fewer parameters than other methods. Furthermore, its effectiveness is sustained even in noisy environments, highlighting its superiority for real-world ECG applications. In the future, we plan to integrate MSECG with various downstream ECG tasks requiring long-term monitoring, such as CA classification, rhythm detection, and respiratory estimation.

\vfill\pagebreak

\bibliographystyle{IEEEtran}
\bibliography{IEEEexample}

\end{document}